%% file: 00_main_review.tex
\documentclass[10pt,twocolumn,letterpaper]{article}

\usepackage{iccv}
\usepackage{times}
\usepackage{epsfig}
\usepackage{graphicx}
\usepackage{amsmath}
\usepackage{amssymb}
\usepackage{multirow}
\usepackage{bm}
\usepackage{booktabs}

\usepackage[pagebackref,breaklinks,colorlinks]{hyperref}
\iccvfinalcopy 

\def\iccvPaperID{3027} 

\ificcvfinal\fi

\def\etal{\emph{et~al}\onedot}

\def\secmk{Sec.~}
\def\figmk{Fig.~}
\def\tablemk{Tab.~}
\def\equationmk{Eqn.~}
\def\supmat{{\color{red}{Sup.~Mat~}}}

\def\ourdata{\emph{Two-hand 500K~}}
\def\Equal{\texttt{=}}

\begin{document}
\title{Reconstructing Interacting Hands with Interaction Prior from Monocular Images}

\author{Binghui Zuo\textsuperscript{1}\hspace{2mm}
Zimeng Zhao\textsuperscript{1}\hspace{2mm}
Wenqian Sun\textsuperscript{1} \hspace{2mm}
Wei Xie\textsuperscript{1}\hspace{2mm}
Zhou Xue\textsuperscript{2} \hspace{2mm}
Yangang Wang\textsuperscript{1}\footnotemark[1]\\%
\\
\textsuperscript{1}Southeast University, China \hspace{2mm}
\textsuperscript{2}Pico IDL, ByteDance, Beijing
}


\maketitle
\ificcvfinal\thispagestyle{empty}\fi

\begin{abstract}
   Reconstructing interacting hands from monocular images is indispensable in AR/VR applications. Most existing solutions rely on the accurate localization of each skeleton joint. However, these methods tend to be unreliable due to the severe occlusion and confusing similarity among adjacent hand parts. This also defies human perception because humans can quickly imitate an interaction pattern without localizing all joints. Our key idea is to first construct a two-hand interaction prior and recast the interaction reconstruction task as the conditional sampling from the prior. To expand more interaction states, a large-scale multimodal dataset with physical plausibility is proposed. Then a VAE is trained to further condense these interaction patterns as latent codes in a prior distribution. When looking for image cues that contribute to interaction prior sampling, we propose the interaction adjacency heatmap (IAH). Compared with a joint-wise heatmap for localization, IAH assigns denser visible features to those invisible joints. Compared with an all-in-one visible heatmap, it provides more fine-grained local interaction information in each interaction region. Finally, the correlations between the extracted features and corresponding interaction codes are linked by the ViT module. Comprehensive evaluations on benchmark datasets have verified the effectiveness of this framework. The code and dataset are publicly available at ~\url{https://github.com/binghui-z/InterPrior_pytorch}. 
\end{abstract}
\renewcommand{\thefootnote}{\fnsymbol{footnote}}
\footnotetext[1]{Corresponding author. E-mail: yangangwang@seu.edu.cn. All the authors from Southeast University are affiliated with the Key Laboratory of Measurement and Control of Complex Systems of Engineering, Ministry of Education, Nanjing, China. This work was supported in part by the National Natural Science Foundation of China (No. 62076061), in part by the Natural Science Foundation of Jiangsu Province (No. BK20220127).}

\input{01_introduction.tex}
\input{02_relatedwork.tex}

\input{03_method.tex}
\input{04_experiments.tex}
\input{05_conclusion.tex}

{\small
\bibliographystyle{ieee_fullname}
\bibliography{egbib}
}
\end{document}

%% file: 01_introduction.tex
\section{Introduction}
\begin{figure}[!t]
    \centering
    \includegraphics[width=\linewidth]{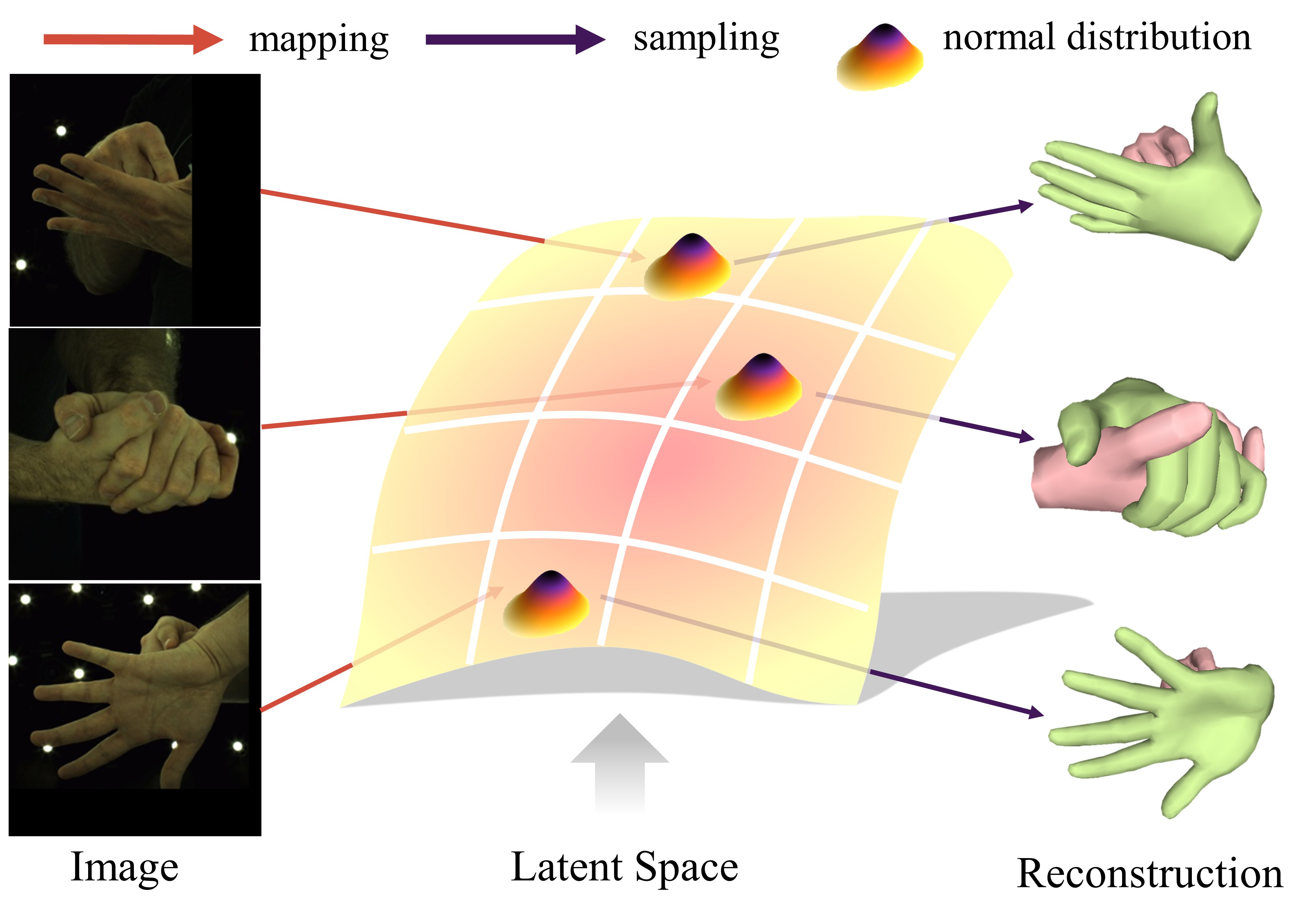}
    \caption{\textbf{Illustration of reconstructing interacting hands from monocular images by our framework}. \emph{Left:} Using a ViT-based fusion network, we map the extracted features from inputs to the learned latent space. \emph{Right:} We sample reasonable reconstructions from the pre-built interaction prior.}
    \label{fig02_teaser}
\end{figure}




Reconstruction of interacting hands is significant for enhancing the behavioral realism of digital avatars in communication, thinking and working. 
With the advent of the RGB dataset~\cite{moon2020interhand2} recording two-hand interactions, numerous attempts have been implemented to reconstruct interacting hands from monocular RGB images. Inspired by the existing single-hand frameworks~\cite{iqbal2018hand,zhou2020monocular,chen2021camera}, pioneer works~\cite{zhang2021interacting,fan2021learning} localize and identify all two-hand joints as the interacting clues. Unfortunately, this process can be seriously misguided by the regional occlusion and local similarity between hands. Subsequent improvements include optimizing re-projection errors~\cite{rong2021monocular}, localizing mesh vertices from coarse to fine~\cite{li2022interacting}, and querying all-in-one visible heatmap~\cite{kim2021end, hampali2022keypoint}. Nevertheless, they still rely on more accurate joint 2D estimators, more diverse marker-less training data, or more computational complexity. 

To overcome this hurdle, our key idea is to \textbf{first construct a comprehensive interaction prior with multimodal datasets and then sample this pre-built prior according to the interaction cues extracted from a monocular image.} It is noted that existing frameworks are always trained with paired data of calibrated images and mesh annotations. This may lead to difficulty in generalizing since the well-known benchmark~\cite{moon2020interhand2} contains simple backgrounds and only around 8.5K interaction patterns. We break this images-paired manner and construct an interaction prior with multimodal datasets, including marker-based data, marker-less data and hands-object data. To do this, a dataset with 500K two-hand patterns is proposed, which contains physically plausible 3D hand joints and MANO parameters. This dataset is used for the unsupervised training of a prior container, which can be formulated by a VAE~\cite{kingma2013auto}. As a result, each two-hand interaction pattern is mapped to an interaction code in the prior space. Since the correlation between the two hands is considered, this representation is more compact than doubling the hand joint/vertex positions or MANO parameters~\cite{romero2017embodied}.

We argue that accurate joint localization is challenging for the monocular reconstruction of interacting hands. As an alternative, we sample the above pre-built interaction prior according to the \emph{interaction adjacency heatmap} (IAH). This heatmap is defined as the mixture coordinate distribution of this joint and other two-hand joints within its coordinate neighborhood. Compared with the 2.5D joint heatmap~\cite{iqbal2018hand, zhang2021interacting}, our IAH abandons the pseudo depth and concatenates more on spatial correlations of the target joint. This heatmap formulation is easier to regress because even for an invisible joint, humans can determine its identity and location according to its spatial neighborhood. Considering that the Gaussian distribution has a more ambiguous boundary, the Laplacian distribution~\cite{laplace1999wiki} is selected as the kernel function of each joint. This effectively reduces the aliasing of interacting adjacency information. These IAHs are further converted to be the corresponding interaction codes through the ViT~\cite{dosovitskiyimage} module and then are regarded as conditions to sample reasonable interaction from the latent space. 

In summary, our main contributions are:

\noindent$\bullet$ A powerful interaction reconstruction framework that compactly represents two-hand patterns as latent codes, which are learned from multimodal datasets in an unsupervised manner. 

\noindent$\bullet$ An effective feature extraction strategy that utilizes interaction adjacency as clues to identify each joint, which is inspired by human perception and is more friendly for network learning. 

\noindent$\bullet$ A large-scale multimodal dataset that records 500K patterns of closely interacting hands, which is more conducive to the construction of our latent prior space. 

%% file: 02_relatedwork.tex
\section{Related Work}
\begin{figure*}[!t]
    \centering
    \includegraphics[width=\linewidth]{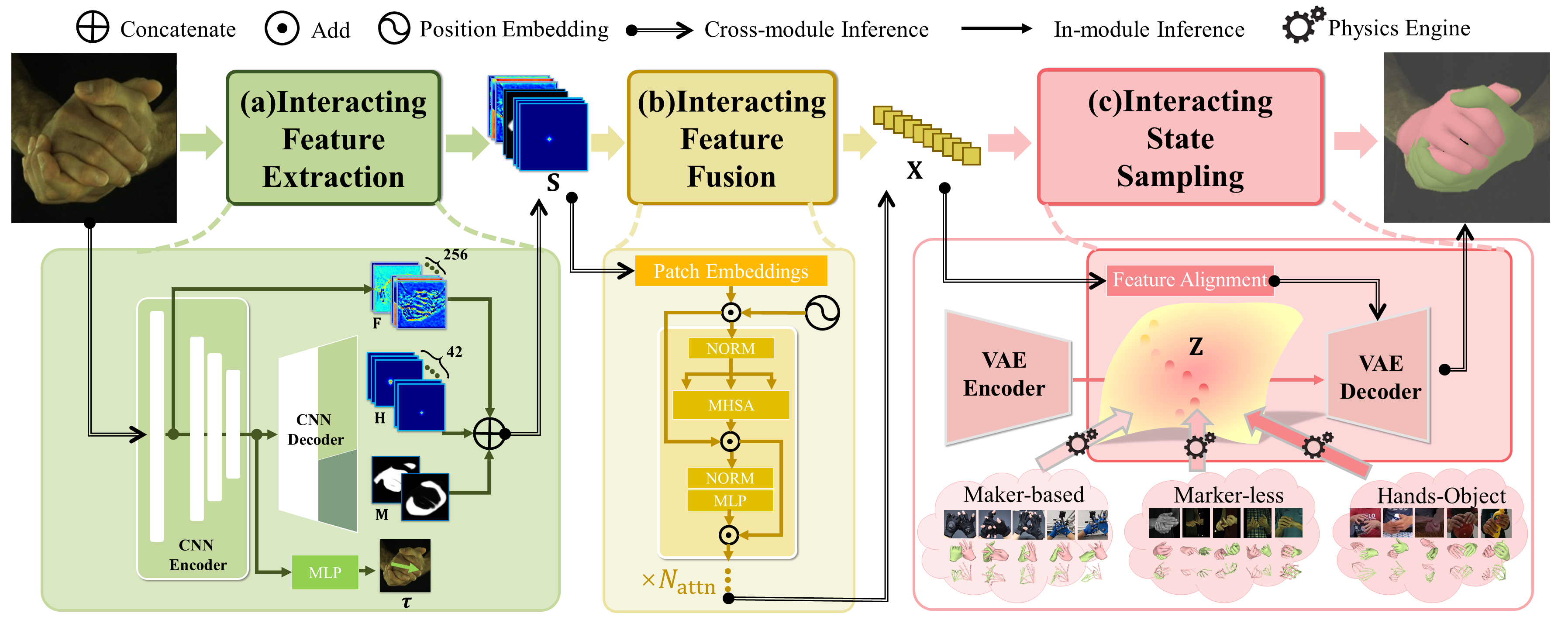}
    \caption{\textbf{Overview of our architecture}. The total pipeline consists of three stages. (a) We first design an expressive feature extraction module to extract global and local context information. Besides that, the proposed IAH adaptively maps adjacent joints with Laplacian distribution and provides more visual attention. (b) A ViT-based fusion module is designed to fuse the extracted features, which are regarded as sampling conditions to reconstruct the final result. (c) We build a powerful interaction prior with multimodal datasets and use the physics engine to make the ground truth more physically plausible.}
    \label{fig03_pipeline}
\end{figure*}

\noindent\textbf{Hand monocular reconstruction} has received a breakthrough after more than a decade of development. Previous studies have always focused on 3D joint estimation~\cite{cai2018weakly, iqbal2018hand, mueller2018ganerated, spurr2020weakly, zimmermann2019freihand, zimmermann2017learning, fan2020adaptive}.~\cite{zimmermann2017learning} introduced the first baseline to predict 3D joint position from a single image. The proposal of MANO~\cite{romero2017embodied} brings a new research direction to this field. With~\cite{boukhayma20193d} presenting the first end-to-end solution for learning 3D hand shape and pose from a monocular RGB image, estimating model parameters~\cite{chen2021model,zhou2020monocular, boukhayma20193d, zhang2019end, chen2021camera, zhang2021hand} from inputs has become a major trend. Some researches~\cite{ge20193d, tang2021towards, kulon2020weakly, wan2020dual} even directly regressed 3D vertices from the inputs. However, most of them are only applicable to a single-pose representation. In this work, we construct an interaction prior which is used to effectively estimate plausible hand poses. Our proposed unified framework can be applied to 3D joints, vertices and MANO parameters.

\noindent\textbf{Interacting hand reconstruction} is critical to promote the development of human-computer interaction(HCI). Due to the severe self-occlusion and the similar appearance of the entangled two hands, previous works heavily relied on depth cameras~\cite{mueller2019real, taylor2017articulated, tzionas2016capturing, oikonomidis2012tracking} or multi-view cameras~\cite{ballan2012motion, han2020megatrack}. Benefiting from the promotion of deep learning and the proposal of interacting hand dataset~\cite{moon2020interhand2}, previous works~\cite{li2022interacting,zhang2021interacting,rong2021monocular,fan2021learning,hampali2022keypoint,kim2021end,meng20223d} tried to estimate interacting hand pose from monocular color images. Most of them attempted to extract distinguishable features of each hand~\cite{kim2021end,fan2021learning,moon2020interhand2,wang2020rgb2hands} or decouple the interaction~\cite{meng20223d,rong2021frankmocap,lin2021two}. Unfortunately, the traditional feature extraction schemes are unsuitable for extracting effective interaction details, and it is unreliable to decouple hands relying on features. Recently, an attention mechanism has been widely adopted~\cite{zhang2021interacting, li2022interacting, hampali2022keypoint} to yield more interacting attention features. Among them, Zhang \etal~\cite{zhang2021interacting} utilized pose-aware attention and context-aware refinement module to improve the pose accuracy. Hampali \etal~\cite{hampali2022keypoint} employed a transformer architecture to model interaction, but the joint angle representation did not perform perfectly. Li \etal~\cite{li2022interacting} fed the bundled features to the attention module, progressively regressing the two-hand vertices. Besides,~\cite{lin2021mesh, lin2021end, cho2022cross} further demonstrated the power of the transformer. Inspired by the above studies, we propose a novel feature extraction module that gives more attention to the interaction region. Meanwhile, a ViT-based model is used to fuse them.

\noindent\textbf{Learning-based prior} has sparked more attention in different domains. Both VAE~\cite{kingma2013auto} and GAN~\cite{Goodfellow2014} are the mainstream models for constructing diverse priors. Most related researchers constructed a prior to avoid implausible situations. To model human motion,~\cite{li2021task, petrovich2021action, ling2020character, kaufmann2020convolutional} used VAE to build motion prior. Other pioneers~\cite{hanocka2020point2mesh, yang2021deep, huang2022pose2uv,huang2022object, pavlakos2019expressive, mittal2022autosdf} introduced pose or shape prior to refine geometric details. Similarly,~\cite{kanazawa2018end, kocabas2020vibe} designed adversarial prior to learn plausible reconstruction. Most similar to us are~\cite{spurr2018cross, wan2017crossing, yang2019aligning, yang2019disentangling}, who applied the built prior to hand pose estimation. Wan \etal~\cite{wan2017crossing} combined GANs and VAEs to build two separate latent spaces. To estimate hand pose from depth maps, they used a mapping function to connect these two spaces. Spurr \etal~\cite{spurr2018cross} proposed a cross-modal framework where both RGB images and depth maps can be used to build the prior. In addition,~\cite{yang2019aligning, yang2019disentangling} committed to aligning the learned latent space with different modalities jointly. 
To this end, we extend the VAE framework to build a novel interaction prior. Compared with competitors, the biggest challenge of two-hand reconstruction is the lack of interaction status. Existing two-hand reconstruction methods are trained on~\cite{moon2020interhand2} with the paired images.  The smaller number of interaction states (8.5K) limits their generalization performance. Fortunately, our core improvement is to construct prior without paired images, which means multimodal datasets can be applied. To compensate for the lack of interaction between two hands, we provide a larger two-hand dataset \ourdata, which has more interaction states than~\cite{moon2020interhand2}.

%% file: 03_method.tex
\section{Method}
The pipeline of the proposed method is shown in \figmk\ref{fig03_pipeline}, which contains three stages: interacting feature extraction (\secmk\ref{sec31_feature_extr}), interacting feature fusion (\secmk\ref{sec32_PosePrior}) and interacting state sampling (\secmk\ref{sec33_reconstruction}). To facilitate the formulation, the hat superscripts represent the predicted result from the network, the star superscripts represent the ground truth values.

\subsection{Interacting Feature Extraction}
\label{sec31_feature_extr}
\begin{figure}[!t]
    \centering
    \includegraphics[width=\linewidth]{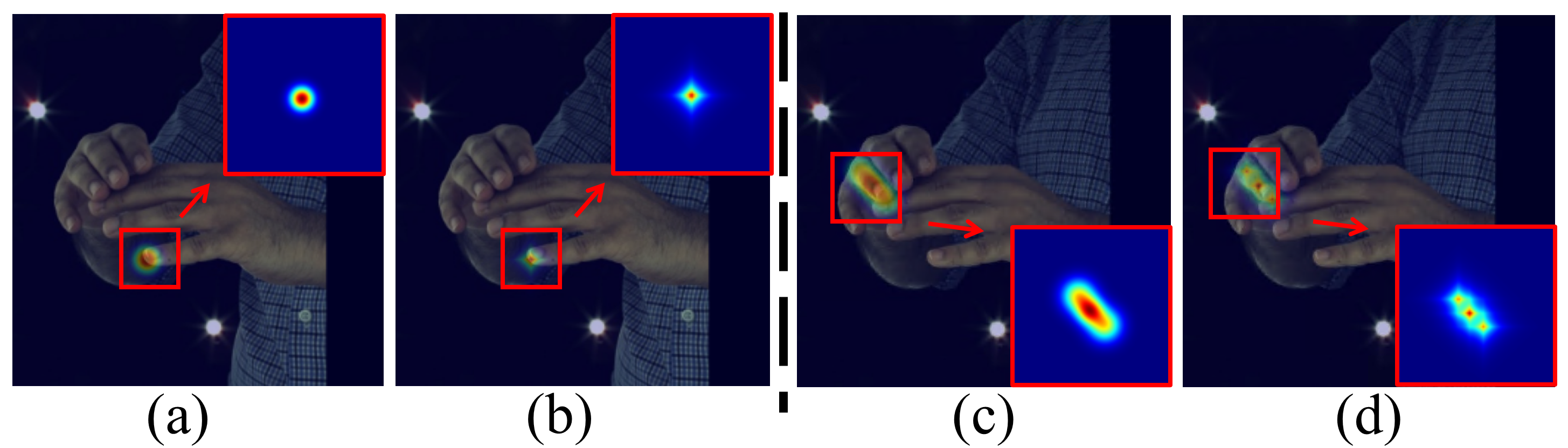}
    \caption{\textbf{IAHs of different channels}. (a) refers to $\mathbf{H}_{42}$ with Gaussian distribution, only the identity joint $\bm{\jmath}_{42}$ is mapped to the heatmap; (b) $\mathbf{H}_{42}$ with Laplacian distribution; (c) refers to $\mathbf{H}_{16}$, both $\bm{\jmath}_{16}$ and adjacent joints are mapped; (d) $\mathbf{H}_{16}$ with Laplacian distribution.} 
    \label{fig02_IAHs}
\end{figure}

\noindent\textbf{Feature selection.}
Given a monocular $\mathbf{I} \in \mathbb{R}^{(3, H, W)}$, we first extract the following 2D features related to the corresponding hand interaction: (i) Interaction adjacency heatmap (IAH) $\{\mathbf{H}_\text{j}\}_{\text{j}=1}^{42}$. (ii) Instance saliency maps $\mathbf{M} \triangleq \mathbf{M}_l\oplus\mathbf{M}_r$. (iii) Left-to-right relative translation $\bm{\tau} \in \mathbb{R}^3$. Because the joints of interacting hands are overcrowded and similar in $\mathbf{I}$, we argue that the strategy of simultaneous identification and localization\cite{iqbal2018hand,zhang2021interacting} is too tough for our task. As an alternative, our IAH emphasizes joint-wise identification more than localization. For a joint located at $\bm{\jmath}_\text{j}(u_\text{j}, v_\text{j})$ and a $d\times d$ image adjacency $\mathcal{A}_\text{j}(d)$, its IAH ground-truth is defined as a 2D Laplacian mixture distribution: 
\begin{equation}
    \begin{aligned}
        \mathbf{H}_\text{j}^*(u, v) &= \frac{1}{2 \sigma_\text{j}} \exp (-\frac{|u-u_\text{j}| + |v-v_\text{j}|}{\sigma_\text{j}}) \\
                               & + \sum_{\bm{\jmath}_\text{k} \in \mathcal{A}_\text{j}(d)} \frac{1}{2 (\alpha \sigma_\text{j}) } \exp (-\frac{|u-u_\text{k}| + |v-v_\text{k}|}{\alpha\sigma_\text{j}})\\
    \end{aligned}
    \label{equation_0}
\end{equation}
where the first term is the Laplacian kernel for the identity $\bm{\jmath}_\text{j}$ with the variance $\sigma_\text{j}$. The other terms are the Laplacian kernels for adjacent joints $\bm{\jmath}_\text{k} \in \mathcal{A}_\text{j}(d)$ with the variance $\alpha \sigma_\text{j}$. $\alpha > 1$ is a zoom factor. As shown in \figmk\ref{fig02_IAHs}(c)\&(d), we select Laplacian instead of Gaussian as the kernel function mainly due to its clearer distribution boundary. ~~$\mathbf{M}$ contains the visible parts of the left and right hand. Because $\bm{\tau}$ is often regarded as a global feature~\cite{zhang2021interacting}, it is regressed by an extra MLP. The overall loss term of this part is: 
\begin{equation}
    \begin{aligned}
        L_{2D}=\sum_{\text{j}=1}^{42}\|\hat{\mathbf{H}}_\text{j} - \mathbf{H}_\text{j}^\star\|_2^2 + \lambda_1\|\hat{\mathbf{M}}- \mathbf{M}^\star\|_2^2 + \lambda_2\|\hat{\bm{\tau}}- \bm{\tau}^{*}\|_2^2
    \end{aligned}
    \label{equation_1}
\end{equation}
In practice, ResNet50~\cite{he2016deep} (with 4 cascading residual blocks) is selected as the feature extraction backbone, and the decoder for 2D local features is designed as a symmetrical structure with 4 blocks. For MLP used for regressing $\bm{\tau}$, after passing through an adaptive pooling layer behind the high-level feature map, we connect two fully connected layers to obtain $\bm{\tau}$.  

\noindent\textbf{Latent utilization.} Besides the above explicitly supervised features, we further utilize the low-level feature maps $\mathbf{F}$ from the first block of our extractor as additional visual guidance. As shown in \figmk\ref{fig03_pipeline} (a), it contains more dense responses and the same map size as $\mathbf{H}$ and $\mathbf{M}$. 

\noindent\textbf{Implementation details.} 
In our experiment, the variance $\sigma_\text{j}$ of identity $\bm{\jmath}_\text{j}$ is set to 2.0, the zoom factor $\alpha$ is set to 2.0 and adjacent region size $d$ is set to 2.5. To balance each loss term, we set $\lambda_1$\Equal$1$ and $\lambda_2$\Equal$2000$. To normalize the translation, we fix the right translation to 0 and predict $\bm{\tau}$ between the left to right hand. ~~We use Pytorch~\cite{Pytorch} to implement the feature extraction network and train it on a single NVIDIA GeForce RTX 3090. To update the network parameters, we use Adam~\cite{kingma2014adam} optimizer with a fixed learning rate 1e-4. We set the batch size to 64 and total training iterations to 500K. Before training, we crop the interacting hand regions with the annotated hand 2D vertices coordinates and resize it to 256$\times$256. To improve the generalization, we perform data augmentation, including random rotation, random flip and color blur.


\subsection{Interacting State Sampling}
\label{sec32_PosePrior}
\noindent\textbf{Prior construction.}
Building the prior allows us to sample the reasonable interaction from the extracted features even if one of the hands is completely occluded. Therefore, the expressiveness and accuracy of the constructed prior directly affect the final reconstruction. Similar to~\cite{spurr2018cross, wan2017crossing, yang2019aligning, yang2019disentangling}, we deploy the VAE framework to build the interaction prior, which consists of an encoder and a decoder. The encoder implicitly maps the input $\bm{x}$ to $p(\bm{z})$ that conforms to the normal distribution, where $p(\bm{z})$ is the prior on the latent space. The decoder reconstructs $\hat{\bm{x}}$ that is close to $\bm{x}$.  We represent the encoder as the conditional probability distribution $q\left(\bm{z} \mid \bm{x}\right)$ and the decoder as $p\left(\hat{\bm{x}} \mid \bm{z}\right)$. The building process is shown in \equationmk\ref{equation_2}.
\begin{equation}
    \begin{aligned}
        \bm{x}\stackrel{q\left(\bm{z} \mid \bm{x}\right)}{\longrightarrow} p(\bm{z}) \stackrel{p\left(\hat{\bm{x}} \mid \bm{z}\right)}{\longrightarrow} \hat{\bm{x}}
    \end{aligned}
    \label{equation_2}
\end{equation}

\noindent\textbf{Training procedure.}
Both the encoder and decoder are composed of fully connected layers that follow ReLU activations. The encoder is a four-layers feed-forward network that models the input $\bm{x}$ to the latent space with dimension $d_z$. We force the output dimension of the encoder to $2d_z$, where the first $d_z$ is used for the regression of mean $\mu$ and the second for variance $\sigma$. 
We strive to shape the distribution with $\mu$ and $\sigma$ into a standard normal distribution and encourage it using Kullback-Leibler divergence loss. Afterward, the latent variable $\bm{z}$ sampled by the reparameterization trick is passed to the decoder to reconstruct the expected result $\hat{\bm{x}}$, where the decoder consists of six linear layers. We use MSE as the reconstruction loss to supervise it. The total loss for interaction prior is defined as:
\begin{equation}
    \begin{aligned}
        L_{\textit{prior}} &= L_{\textit{KL}}(\bm{z}) + \lambda_3 \|\hat{\bm{x}} - \bm{x}\|_2^2
    \end{aligned}
    \label{equation_3}
\end{equation}
In the inference phase, we discard the encoder and only use the decoder with the frozen parameters to obtain the reconstruction. 

\noindent\textbf{Hand representation.}
Benefiting from the embedding ability of VAE, we conduct experiments on three different hand pose representations, including 3D joints coordinates, 3D vertices coordinates and MANO parameters~\cite{romero2017embodied}. For compatibility with our framework, only pose and shape parameters are considered when embedding MANO parameters, as the relative translation has been estimated in the feature extraction module.

\noindent\textbf{Implementation details.} 
To balance quality and generalization, we set loss weight $\lambda_3$ to $100$ to ensure losses are within one order of magnitude~\cite{ling2020character}. During training, we flatten inputs as a vector and map them to latent space with $d_z$\Equal128. We use Adam~\cite{kingma2014adam} optimizer with a base learning rate of 1e-5 and a batch size of 64. 

\subsection{Interacting Feature Fusion}
\label{sec33_reconstruction}
\noindent\textbf{Feature fusion.}
To effectively leverage the features extracted in \secmk\ref{sec31_feature_extr}, a ViT-based fusion~\cite{dosovitskiyimage} module with powerful attention is used to fuse them. Specifically, the global image feature $\mathbf{F}$, interaction adjacency heatmap $\mathbf{H}$ and instance saliency maps $\mathbf{M}$ are concatenated to form fusion features $\mathbf{S} \in \mathbb{R}^{(C, H, W)}$. All of them have the same resolution $H$\Equal$W$\Equal$64$, and the channel $C$ of $\mathbf{S}$ is 300. Before ViT, we reshape $\mathbf{S}$ into $n$ visual patches,  $n$\Equal$(HW/P^2)$, $P$ denotes the size of each patch. We also use a single linear layer to map the patches into patch embeddings $\mathbf{S}^{\prime}$ with dimension $D$, which is equal to the hidden size of all the transformer layers~\cite{dosovitskiyimage}. Extra position embeddings for preserving spatial information are added to the patch embeddings $\mathbf{S}^{\prime}$. The fusion network follows a standard Transformer structure, which consists of multi-head self-attention layers $(\operatorname{ATTE})$, feed-forward blocks $(\operatorname{FF})$ and normalization layers $(\operatorname{LN})$. We feed the embeddings into $N_{attn}$ transformer attention blocks according to the following equation.
\begin{equation}
    \begin{aligned}
        \tilde{\mathbf{S}}_{n+1}^{\prime}=\mathbf{S}_n^{\prime}+\operatorname{ATTE}\left(\operatorname{LN}\left(\mathbf{S}_n^{\prime}\right)\right), \\ 
        \quad \mathbf{S}_{n+1}^{\prime}=\tilde{\mathbf{S}}_{n+1}^{\prime}+\operatorname{FF}\left(\operatorname{LN}\left(\tilde{\mathbf{S}}_{n+1}^{\prime}\right)\right)
    \end{aligned}
    \label{equation_4}
\end{equation}
where $n$ denotes different transformer blocks and the tilde superscripts represent the intermediate output of the transformer.

\noindent\textbf{Feature alignment.}
To ensure the consistency between the dimension of ViT output and the pre-built interaction prior, we add a feature alignment block consisting of a linear layer after the final transformer block. We treat the output of the feature alignment block as a condition and sample the expected reconstruction from the pre-built interaction prior. Only the VAE decoder with frozen parameters is employed for this process.
\begin{figure}[!t]
    \centering
    \includegraphics[width=\linewidth]{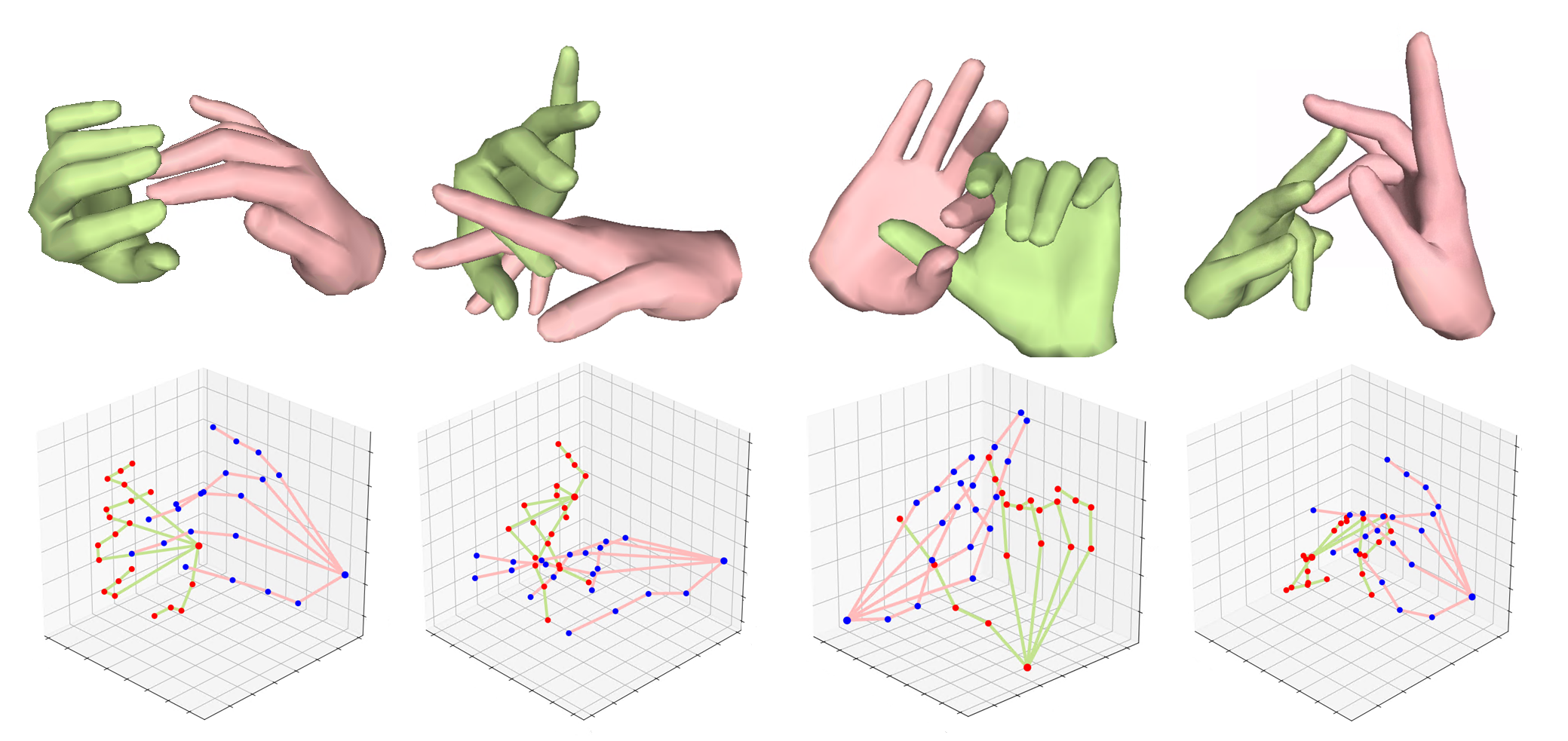}
    \caption{\textbf{The proposed \ourdata}. Both 3D joint positions and corresponding MANO parameters are provided.}
    \label{fig0_dataset}
\end{figure}

\noindent\textbf{Training procedure.}
We train the ViT-based fusion and simultaneously fine-tune the feature extraction module in an end-to-end manner. In addition to the feature extraction loss defined in \equationmk\ref{equation_1}, we also apply other special losses to supervise the reconstruction of different hand representations. For the representation of 3D joints and 3D vertices, we use the $L_1$ distance as a loss function to ensure consistency between the prediction and ground truth. And for the representation of MANO parameters, we also adopt additional loss terms to make the hand surface smoother and physically plausible, including normal loss~\cite{li2022interacting,moon2020i2l} and penetration loss~\cite{jiang2020coherent, rong2021monocular}. 

\noindent\textbf{Implementation details.} 
In our framework, we set the size of each patch $P$ to 8, patch embeddings size $D$ to 1024 and total use $N_{attn}$\Equal$6$ transformer blocks. Different from feature extraction and interaction prior module, both the Adam~\cite{kingma2014adam} optimizer and the learning rate scheduler are used. The training process costs 150 epochs with a batch size of 64 on a single NVIDIA GeForce RTX 3090. We fix the learning rate at 1e-4 in the first 100 epochs and then adaptively reduce it with the scheduler. Besides the above, as the parameters of the feature extraction module are also updated, we use the same data augmentation as the training of the feature extraction network. 

\subsection{Interacting Modality Expansion}
\noindent\textbf{Motivation.} To eliminate the dependence on image datasets when building the interaction prior, we propose the following measures to increase the diversity of hand interaction: (i) Capturing skeleton data according to the marker-based system simultaneously; (ii) Randomly combining the poses sampled from single-hand datasets. 
\figmk\ref{fig0_dataset} shows our \ourdata.

\begin{figure}[!t]
    \centering
    \includegraphics[width=\linewidth]{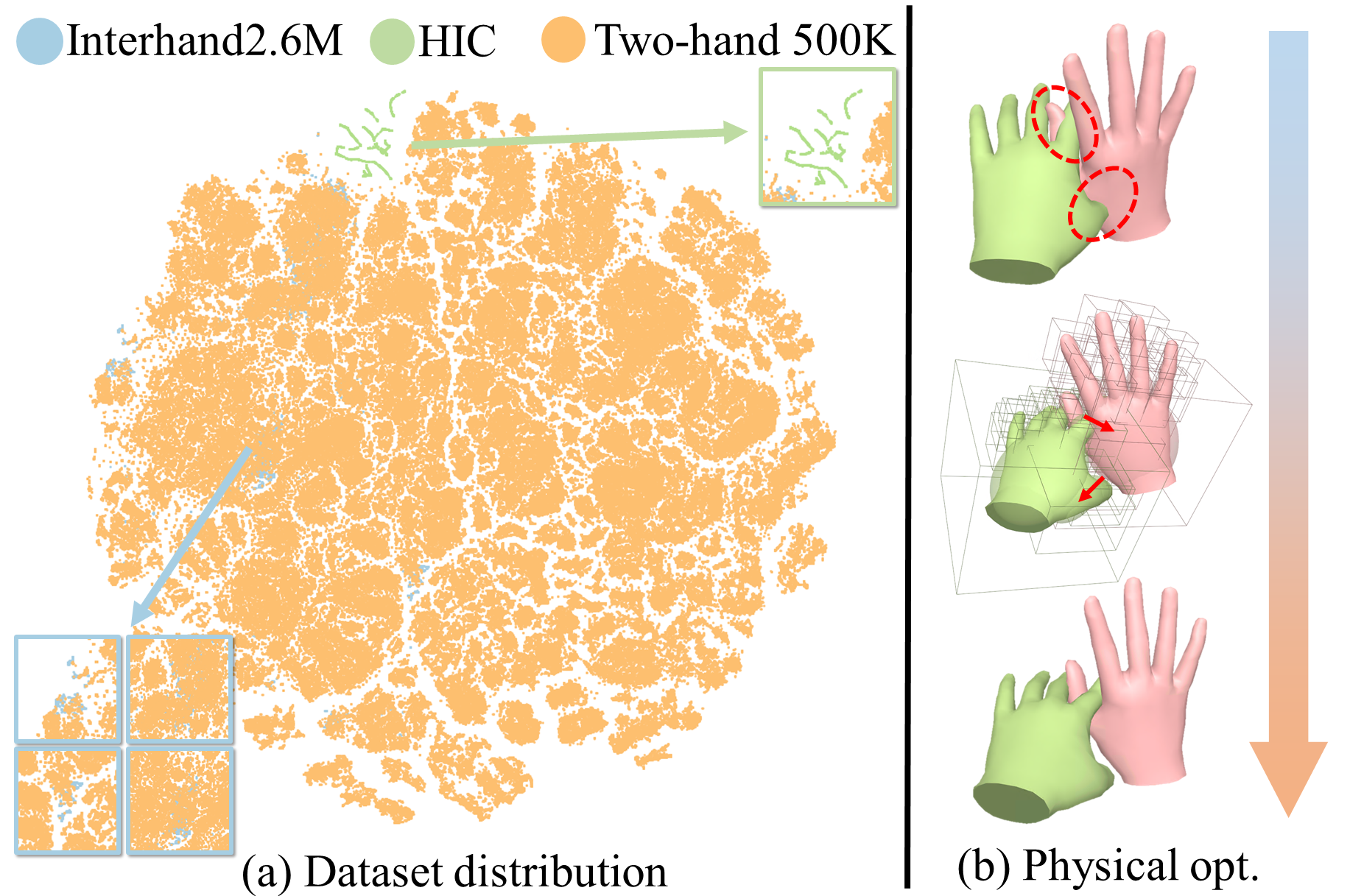}
    \caption{\textbf{Dataset distribution and physical optimization}. (a) t-SNE distribution of~\cite{moon2020interhand2,tzionas2016capturing} and our \ourdata; (b) physical optimization for the plausible interaction. From top to bottom: before optimization, during optimization, after optimization.}
    \label{fig0_tsne}
\end{figure}

\noindent\textbf{More diversity.} To obtain more diverse interaction states, we construct \ourdata in a multimodal manner. With the above data generation measures, more than 500K interaction states are captured. Besides the data captured by the marker-based MoCap system, we splice left-right hand instances sampled from single-hand datasets~\cite{moon2020interhand2, zimmermann2017learning, gomez2019large, zimmermann2019freihand, zhang2017hand}. It should be noted that although sufficient MoCap data could be collected, considering the cost of the MoCap system, it is still meaningful to utilize single-hand data. \figmk\ref{fig0_tsne} (a) visualizes the distribution of related two hand datasets~\cite{moon2020interhand2,tzionas2016capturing} and \ourdata, showing that our proposed dataset is more diverse than them. 

\noindent\textbf{Less penetration.} For the marker-based data, we fit MANO parameters~\cite{romero2017embodied} from 3D skeleton by solving inverse kinematics (IK). As the fingers are often tangled together, penetration and dysmorphism always exist between two hands. To ensure physical interaction, we use the physics engine~\cite{Bullet} to optimize the fitted hand pose. Similar to~\cite{zhao2022stability,huang2022neural}, we adopt a sampling-based optimization scheme to iteratively refine interaction. The same strategy is also applied to the splicing process to ensure the assemblies are plausible. \figmk\ref{fig0_tsne} (b) demonstrates the interaction before and after optimization. For more details about our dataset, please refer to \supmat.

%% file: 04_experiments.tex
\section{Experiments}
\subsection{Datasets and Metrics}
\label{sec42_data} 
\noindent\textbf{Prior data.}
Data from three different domains are used to construct interaction prior in our practice: marker-based data (\ourdata), marker-less data~\cite{moon2020interhand2} and hands-object data~\cite{hampali2022keypoint, kwon2021h2o}. Since the relative translation is estimated by the network, all two-hand data in~\cite{hampali2022keypoint, kwon2021h2o} can be used to construct interaction prior, even if the hands are not strictly interacting. We use the physics engine to process implausible interactions in the dataset.

\noindent\textbf{Reconstruction data.}
We only use~\cite{moon2020interhand2} to train the procedure from image to reconstruction. Before training, we pick out interacting instances and corresponding labels annotated by human and machine (H+M), containing 366K training instances and 261K testing instances. Interacting subjects in~\cite{tzionas2016capturing} are only employed to demonstrate qualitative performance.

\begin{figure}[!t]
    \centering
    \includegraphics[width=\linewidth]{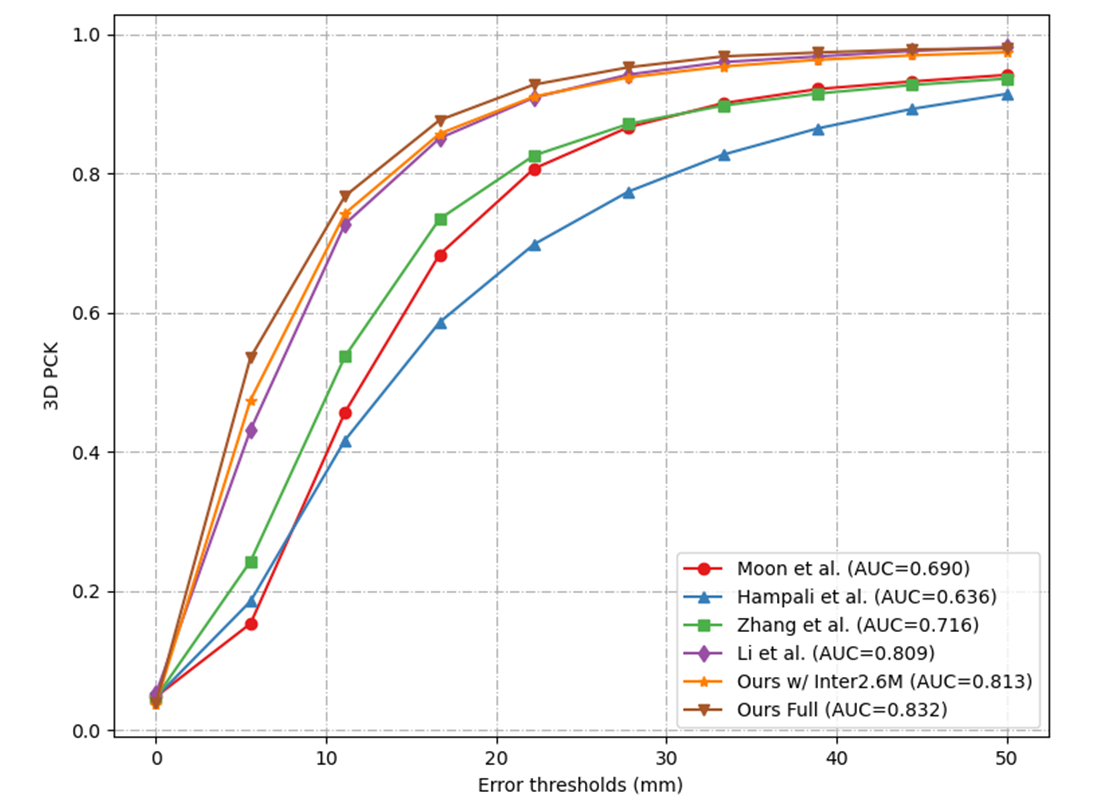}
    \caption{\textbf{Comparisons with the SOTA methods on \emph{Interhand2.6M} dataset}. We construct interaction prior using \emph{Interhand2.6M} for fair comparisons. Among them, Moon \etal refer to~\cite{moon2020interhand2}, Hampali \etal refer to~\cite{hampali2022keypoint}, Zhang \etal refer to~\cite{zhang2021interacting} and Li \etal refer to~\cite{li2022interacting}}
    \label{fig07_PCK}
\end{figure}

\noindent\textbf{Metrics.} 
To evaluate the accuracy of hand pose, we use the mean per joint position error (\emph{MPJPE}) in millimeters. For fair comparisons, we calculate the \emph{MPJPE} after aligning the root joints and scaling the bone lengths of each hand~\cite{zhang2021interacting,li2022interacting,hampali2022keypoint,moon2020interhand2}. Apart from that, the percentage of correct keypoints (\emph{PCK}) and the area under the curve (\emph{AUC}) in the range of 0 to 50 millimeters are taken to assess the evaluations. We also report the mean per vertex position error (\emph{MPVPE}) to evaluate the quality of the reconstructed hand surface.

\subsection{Comparison with the SOTA}
\label{sec43_Comparison}
We use the hand pose representation of MANO parameters to report the comparison results. In fairness, the following comparisons are performed on the basis of constructing interaction prior with only the (ALL) branch of~\cite{moon2020interhand2}. 

\begin{figure*}[!t]
    \centering
    \includegraphics[width=\linewidth]{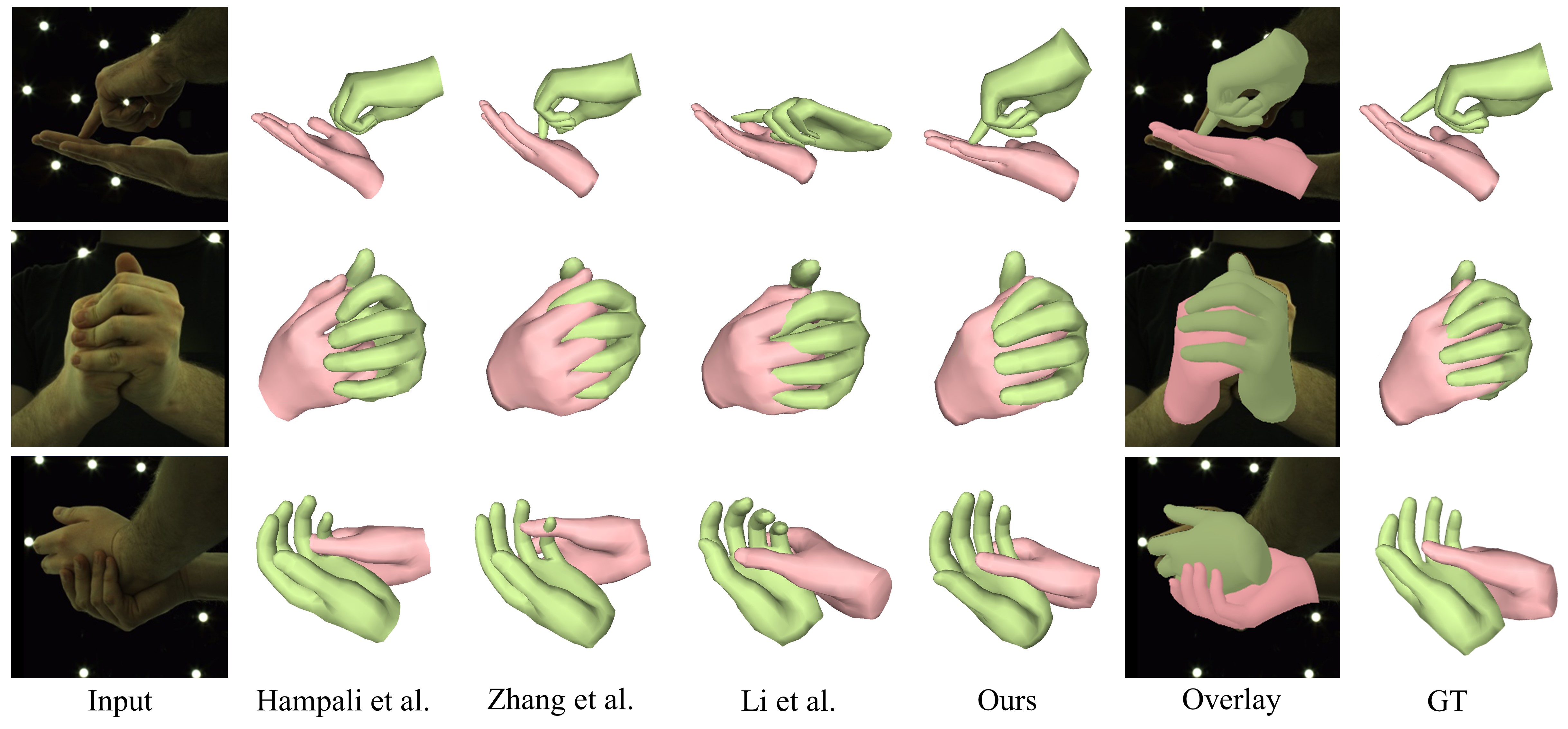} 
    \caption{\textbf{Qualitative results on~\cite{moon2020interhand2}}. Qualitative comparisons with other SOTA methods, including Hampali \etal~\cite{hampali2022keypoint}, Zhang \etal~\cite{zhang2021interacting} and Li \etal~\cite{li2022interacting}, these comparisons demonstrate our method gains more accurate and high-fidelity reconstruction results.}
    \label{fig05_qualitative_res0}
\end{figure*}

\begin{figure*}[!t]
    \centering
    \includegraphics[width=\linewidth]{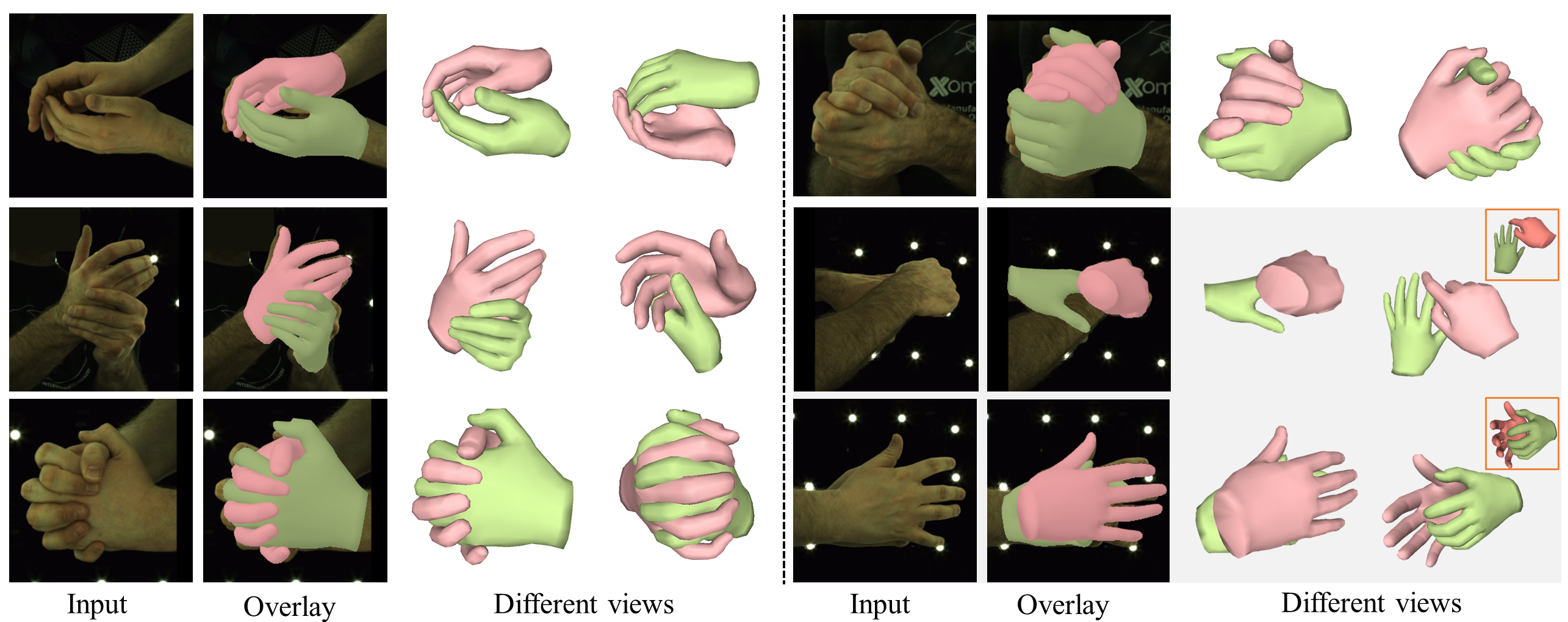}
    \caption{\textbf{More reconstruction results on~\cite{moon2020interhand2}}. High-quality reconstruction results certify the effectiveness of our method. Benefiting from the constructed interaction prior, even with complete self-occlusion (shown in the last two instances), the performance is still satisfactory. The related ground truth is depicted on the top right.}
    \vspace{-2mm}
    \label{fig06_qualitative_res1}
\end{figure*}

\noindent\textbf{Qualitative results}.
Comprehensive comparisons show the satisfactory performance of our method. \figmk\ref{fig05_qualitative_res0} demonstrates the qualitative results compared with previous SOTA methods~\cite{hampali2022keypoint, zhang2021interacting, li2022interacting} for interacting hands reconstruction. Compared to them, our reconstructed interaction generates less penetration while ensuring fidelity, which means that the constructed interaction prior effectively addresses the problem caused by severe occlusion and homogeneous appearance. \figmk\ref{fig06_qualitative_res1} further demonstrates the qualitative analysis of the reconstruction. For the completely occluded instances in the gray background, the reconstructed interaction is also consistent with the ground truth (shown on the top right). More results on~\cite{moon2020interhand2} and~\cite{tzionas2016capturing} can be seen in \supmat.

\noindent\textbf{Quantitative results}.
The summarized results shown in \tablemk\ref{tab03_querymodes} indicate the superior performance of our method. From the first four rows in \tablemk\ref{tab03_querymodes}, we present the reconstruction performance of single-hand reconstruction methods~\cite{zimmermann2017learning, zhou2020monocular, boukhayma20193d, spurr2018cross}. The poor performance suggests that applying the single-hand reconstruction method directly to our task is undesirable due to heavy self-occlusion and appearance confusion. We further compare with almost all recent two-hand reconstruction methods~\cite{moon2020interhand2,fan2021learning,zhang2021interacting,kim2021end,li2022interacting,hampali2022keypoint} in the community. In exception to them,~\cite{meng20223d} is not considered because they actually estimate single hand by interacting hand de-occlusion and removal. Compared to our full model, the interaction prior corresponding to w/\emph{Inter2.6M} is only trained with~\cite{moon2020interhand2}. It can be concluded that non-image paired multimodal training data is positive for constructing interaction prior. \figmk\ref{fig07_PCK} depicts the \emph{PCK} curve of our method, which is also superior to other methods.

\begin{table}[!t]
    \begin{center}
        \resizebox{1\linewidth}{!}{
          \begin{tabular}{c|ccc}
                \noalign{\hrule height 1.5pt}
                Methods.       & MPJPE(mm)$\downarrow$    & MPVPE(mm)$\downarrow$    \\
                \noalign{\hrule height 1pt}
                Zimmermann \etal~\cite{zimmermann2017learning}  & 36.36     & -.               \\
                Zhou \etal~\cite{zhou2020monocular}     & 23.48     & 23.89                   \\
                Boukhayma \etal~\cite{boukhayma20193d}     & 16.93     & 17.98                   \\
                Spurr \etal~\cite{spurr2018cross}       & 15.40     & -                       \\
                \hline
                Moon \etal~\cite{moon2020interhand2}     & 16.53    & -.                   \\
                Hampali \etal~\cite{hampali2022keypoint}    & 20.54    &22.27                \\
                Fan \etal~\cite{fan2021learning}    & 15.37    &-.           \\
                Zhang \etal~\cite{zhang2021interacting}   & 13.77     &14.40                \\
                Kim \etal~\cite{kim2021end}     & 12.52    &-.           \\
                Li \etal~\cite{li2022interacting}         & 8.67      &8.92                 \\
                \hline
                Ours w/ \emph{Inter2.6M}~\cite{moon2020interhand2}       & 8.77      &8.81              \\
                Ours                                   & $\textbf{8.34}$     &$\textbf{8.51}$            \\
                \noalign{\hrule height 1.5pt}
          \end{tabular}
        }
    \end{center}
    \caption{\textbf{Quantitative comparisons on~\cite{moon2020interhand2}}. Row1-Row4 report pose accuracy with single-hand methods. These results are taken from~\cite{li2022interacting,zhang2021interacting}. The following rows report two-hand methods. Among them, we use MANO
    joint angles representation to evaluate~\cite{hampali2022keypoint}.}
    \label{tab03_querymodes}
  \end{table}

\subsection{Ablation Study}
\label{sec44_ablation}
We use the hand pose representation of MANO parameters to ablate the effectiveness of each component, but also report the accuracy of other pose representations with the same configuration. 

\begin{figure}[!t]
    \centering
    \includegraphics[width=\linewidth]{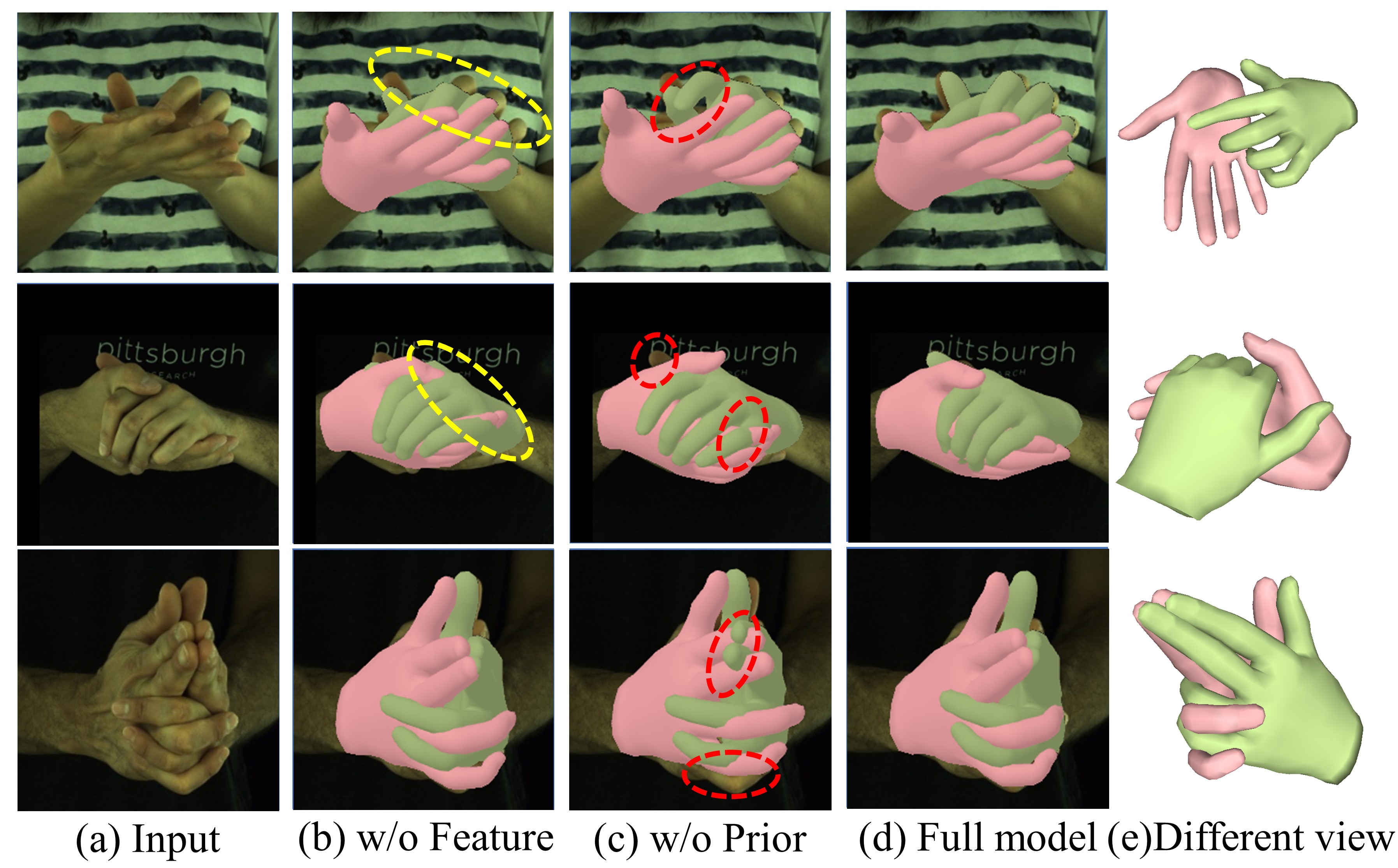}
    \caption{\textbf{Effectiveness of different components}. From left to right: The input images, removing feature extraction module, removing constructed interaction prior, reconstructed interaction with our full model, alternative views.}
    \vspace{-2mm}
    \label{fig07_ablation}
\end{figure}

\noindent\textbf{Effectiveness of feature extraction}.
When extracting interacting features from inputs, we do not overemphasize extracting the unique features for each hand, as the self-occlusion between interacting hands makes it difficult. We design a novel feature extraction module that reflects more global-local context information and report the effectiveness in \tablemk\ref{tab02_ablation}. We first perform ablation by removing the extracted local features and only using the global features $\mathbf{F}$ to reflect all information. The poor performance in Row.a shows the importance of the feature extraction module, indicating that the extracted features provide more interacting clues. We further ablate the impact of each part in the feature extraction module (Row.b and Row.c) by discarding the corresponding part separately. The visualization about the effectiveness of feature extraction is shown in \figmk\ref{fig07_ablation} (b). 

\noindent\textbf{Effectiveness of IAH}.
To demonstrate that IAH is more suitable for interacting hands reconstruction, we express the heatmaps with different forms and list the corresponding effects. As shown in Row.d and Row.e of \tablemk\ref{tab02_ablation}, although the conventional heatmaps help to reduce the errors, the impact on reconstruction is more marginal than our proposed IAH. We attribute this success to the adaptability of IAH. Row.f investigates the IAH with Gaussian distribution, and the inferior performance suggests that the Laplacian distribution is more suitable for IAH. More ablations can be found in \supmat.

\noindent\textbf{Effectiveness of prior.}
Benefiting from the constructed interaction prior, unreal interaction states have been excluded. We analyze the impact of the interaction prior by replacing it with an MLP architecture~\cite{rong2021monocular}. Row.g in \tablemk\ref{tab02_ablation} shows the result without interaction prior. The unsatisfactory performance highlights the importance of interaction prior, denoting it contributes more to improving accuracy than the feature extraction module. Besides, \figmk\ref{fig07_ablation} (c) further demonstrates the significant decrease in mesh quality. 

\noindent\textbf{Effectiveness of ViT-based fusion.}
The effects of both the extracted features and pre-built prior have been analyzed. Maximizing the performance of extracted features and accurately sampling the constructed prior are critical to the final reconstruction. We compare two other powerful backbones ResNet50~\cite{he2016deep} and HRNet32~\cite{sun2019deep}, each of which is applied to completely replace ViT. Comparing Row.h and Row.i in \tablemk\ref{tab02_ablation}, we see that the ViT-based network gives more powers when fusing interaction features. That is because ViT obtains more global-local context information and effectively models the interactions between two hands. It is noted that only 6 ViT blocks are adopted in the experiments, making the parameter count comparable among different networks and ensuring the fairness of the ablations.

\noindent\textbf{Influence of different representations.}
We further compare the interaction priors constructed separately with 3D hand joints, 3D hand vertices and MANO parameters within the same dimension. The corresponding performance is reported in Row.j and Row.k of \tablemk\ref{tab02_ablation}. Among them, the best performance is achieved by the MANO representation, while the lowest accuracy occurs in the representation of 3D vertices. We attribute the reason to the self-restriction of MANO parameters and it is more difficult to embed discrete 3D coordinates.

\noindent\textbf{Discussion on prior structure.}
We discuss two candidate prior structures: auto-encoder (AE) and variational auto-encoder (VAE). AE is data-dependent and can not generate data. While VAE drives the latent variable to conform to the standard normal distribution and uses reparameterization tricks to improve generativity and robustness~\cite{meng2019vv}, which is more reasonable to construct interaction prior.
\begin{table}[!t]
    \begin{center}
        \resizebox{1\linewidth}{!}{
            \begin{tabular}{l|c|l| cc}
                \noalign{\hrule height 1.5pt}
                Comp. &Row. & Variants.                  & MPJPE $\downarrow$    & MPVPE $\downarrow$    \\
                \noalign{\hrule height 1pt}
                \multirow{3}{*}{ 2D Feat.}  & a & w/o feature          & 10.19     & 10.25        \\
                \multirow{3}{*}{}        & b           &   w/o saliency         & 9.06     & 9.11         \\
                \multirow{3}{*}{}        & c           &    w/o IAH            & 9.87     & 10.07         \\
                \hline
                \multirow{3}{*}{Hm. Repr.}    & d               &    w/ an all-in-one            & 9.63     &9.90        \\
                \multirow{3}{*}{}         & e          &   w/ a joint-wise           & 8.81    & 9.05         \\
                \multirow{3}{*}{}         & f          &  w/ Gaussian             & 9.22     & 9.69         \\
                \hline
                Prior   & g       &w/o prior          & 11.07     & 11.49         \\
                \hline
                \multirow{2}{*}{Fusion} & h  &  w/ ResNet50          &  9.22    &  9.27         \\
                \multirow{2}{*}{}           & i        &   w/ HRNet32          &9.16   & 9.20       \\
                \hline
                \multirow{2}{*}{Pose Repr.}  & j &  w/ 3D joints         &  9.74   &  -.      \\
                \multirow{2}{*}{}        & k           &   w/ 3D vertices          & -.  & 10.52      \\
                \hline
                                          & l              &ours   & 8.34     &8.51      \\
                \noalign{\hrule height 1.5pt}
            \end{tabular}
        }
    \end{center}
    \caption{\textbf{Ablation study of components in our framework}. Adequate ablation experiments have explored the effectiveness of each key component: feature extraction module, IAH, interaction prior, feature fusion network and different representations.}
    \vspace{-2mm}
    \label{tab02_ablation}
\end{table}

%% file: 05_conclusion.tex
\section{Conclusion}
This work treats the interacting hands as a whole, constructs interaction prior based on multimodal datasets, and utilizes joint-wise interaction adjacency to reconstruct interacting hands from monocular images. Compared to most existing works, our framework elegantly combines multimodal datasets to build interaction prior and further recasts the reconstruction as the conditional sampling from this prior.
To facilitate its training, \ourdata dataset is further constructed with modal diversity and physical plausibility considerations. Our framework based on cross-modal interaction prior would also bring inspiration to other multi-body reconstruction tasks. 

\noindent\textbf{Limitations and Future Work.}
Although we can obtain reasonable interaction from the constructed interaction prior, the penetration is still unavoidable for complicated entanglements. In the future, using the physics engine to guide interaction could bring more benefits to the community.